\documentclass{llncs}

\usepackage[utf8]{inputenc}
\usepackage[english]{babel}
\usepackage{amsmath}
\usepackage{amsfonts}
\usepackage{graphicx}
\usepackage{xcolor,colortbl}
\usepackage{subfigure}

\begin{document}

\title{Demystifying  Relational  Latent Representations}
%
%
\author{Sebastijan Dumančić \and Hendrik Blockeel}
\authorrunning{Dumančić and Blockeel} 
%
\tocauthor{Sebastijan Duman\v{c}i\'{c} and Hendrik Blockeel}
\institute{Department of Computer Science, KU Leuven Belgium\\
\email{\{firstname.lastname\}@cs.kuleuven.be}}

\maketitle              

\begin{abstract}
Latent features learned by deep learning approaches have proven to be a powerful tool for machine learning.
They serve as a data abstraction that makes learning easier by capturing regularities in data explicitly.
Their benefits motivated their adaptation to relational learning context.
In our previous work, we introduce an approach that learns \textit{relational} latent features by means of clustering instances and their relations.
The major drawback of latent representations is that they are often \textit{black-box} and difficult to interpret.
This work addresses these issues and shows that (1) latent features created by clustering are interpretable and capture interesting properties of data; (2) they identify local regions of instances that match well with the label, which partially explains their benefit; 
and (3) although the number of latent features generated by this approach is large, often many of them are highly redundant and can be removed without hurting performance much.
\end{abstract}

\section{Introduction}

Latent representations created by deep learning approaches \cite{Goodfellow2016} have proven to be a powerful tool in machine learning.
Traditional machine learning algorithms learn a function that directly maps data to the target concept.
In contrast, deep learning creates several layers of latent features between the original data and the target concept.
This results in a multi-step procedure that simplifies a given task before solving it.

The progress in learning such latent representations has predominantly focused on vectorized data representations.
Likewise, their utility has been recognized in the relational learning community \cite{Nickel0TG16} in which models are learned not only from instances but from their relationships as well \cite{Getoor2007,Muggleton94}. 
The prevalent latent representations paradigm in that direction are \textit{embeddings to vector spaces} \cite{Bordes2011,Nickel2011,Bordes2013}.
The core idea behind the embeddings is to replace symbols with numbers and logical reasoning with algebra.
More precisely, relational entities are transformed to low-dimensional vectors and relations to matrices or functions of vectors.
This way of learning latent features corresponds to learning the low-dimensional representations of relational entities and relations.
Many variations of this formalization exist, but they share the same underlying principle.
Assuming facts \texttt{p(a,b)} and \texttt{r(b,a)}, \texttt{a} and \texttt{b} are entities whereas \texttt{p} and \texttt{r} are existing relations between them.
The goal is to find corresponding vectorized representations of \texttt{a} and \texttt{b}, $a$ and $b$ respectively, together with matrix representations of \texttt{p} and \texttt{r}, $P$ and $R$ respectively.
More precisely, the goal is to find vectorized representations such that products $aPb$ and $bRa$ have \textit{high values}.
In contrast, given a false fact \texttt{q(a,b)} product $aQb$ should have a \textit{low value}.

These embeddings approaches have several drawbacks.
First, the latent features created that way have no inherent meaning --  they are created to satisfy the aforementioned criteria.
This is thus a major obstacle for interpretability of the approach, which is important in many aspects and one of the strengths of relational learning.
Second, huge amounts of data are needed in order to extract useful latent features.
Knowledge bases used for training often contain millions of facts.
Third, it is not clear how these approaches can handle unseen entities (i.e., an entity not present in the training set and whose embedding is therefore not known) without re-training the entire model.

Recently, Dumančić and Blockeel \cite{Dumancic2017} introduced a complementary approach, titled CUR$^2$LED, that takes a relational learning stance and focuses on learning relational latent representations in an unsupervised manner.
Viewing relational data as a hypergraph in which instances form vertices and relationships among them form hyperedges, the authors rely on clustering to obtain latent features.
The core component in this approach is a declarative and intuitive specification of the similarity measure used to cluster both instances and their relationships.
This consequently makes entire approach more \textit{transparent} with respect to the meaning of latent features, as the intuitive meaning of similarity is precisely specified.

The benefits of latent representations were clearly shown with respect to both performance and complexity.
The complexity of models learned on latent features was consistently lower compared to the models learned on the original data representation.
Moreover, the models learned with latent features often resulted in improved performance, by a large margin as well.
These two results jointly show that latent representations capture more complex dependencies in a simple manner.

In this work we further investigate the properties of relational latent representations created by CUR$^2$LED.
We start by asking the question: \textit{what do latent features mean?}
We introduce a simple method to extract the meaning of the latent features, and show that they capture interesting properties.
We ask next: \textit{what makes latent representations effective?}
The initial work showed the benefits of the latent representations, however, no explanation is offered why that is the case.
We hope to shed light behind the scene and offer (at least a partial) answer why that is the case.

In the following section we first briefly introduce neighbourhood trees -- a central concept of CUR$^2$LED.
We then describe an approach used in extracting the knowledge form the latent features, and investigating the properties of such latent representation.
The results are presented and discussed next, followed by the conclusion.

\section{Neighbourhood trees}

\begin{figure}[t]
	\centering
    \begin{minipage}{0.48\textwidth}
    	\centering
    	\includegraphics[scale=0.25]{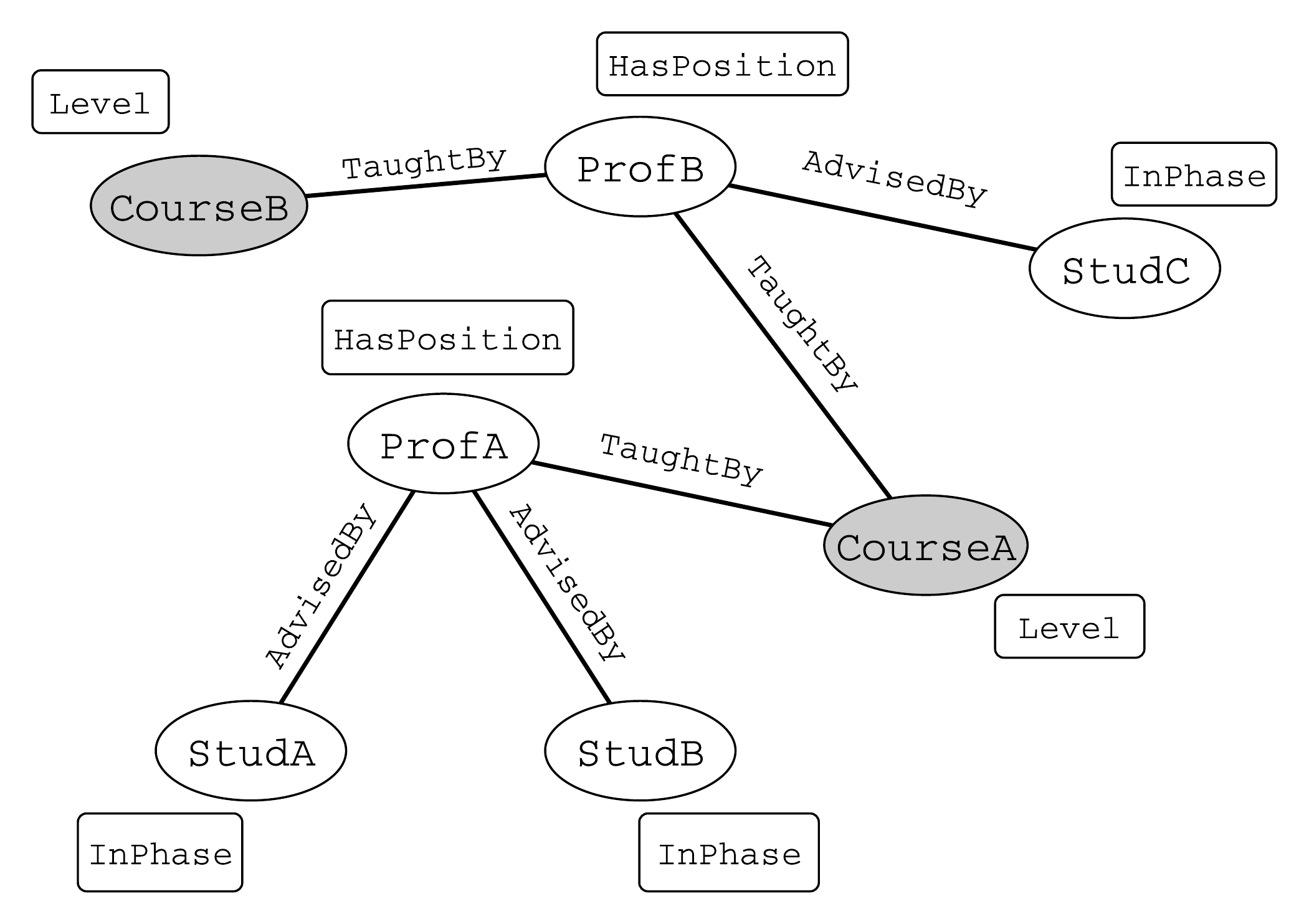}
    \end{minipage}
    \hspace{0.5em}
    \begin{minipage}{0.48\textwidth}
    	\centering
    	\includegraphics[scale=0.25]{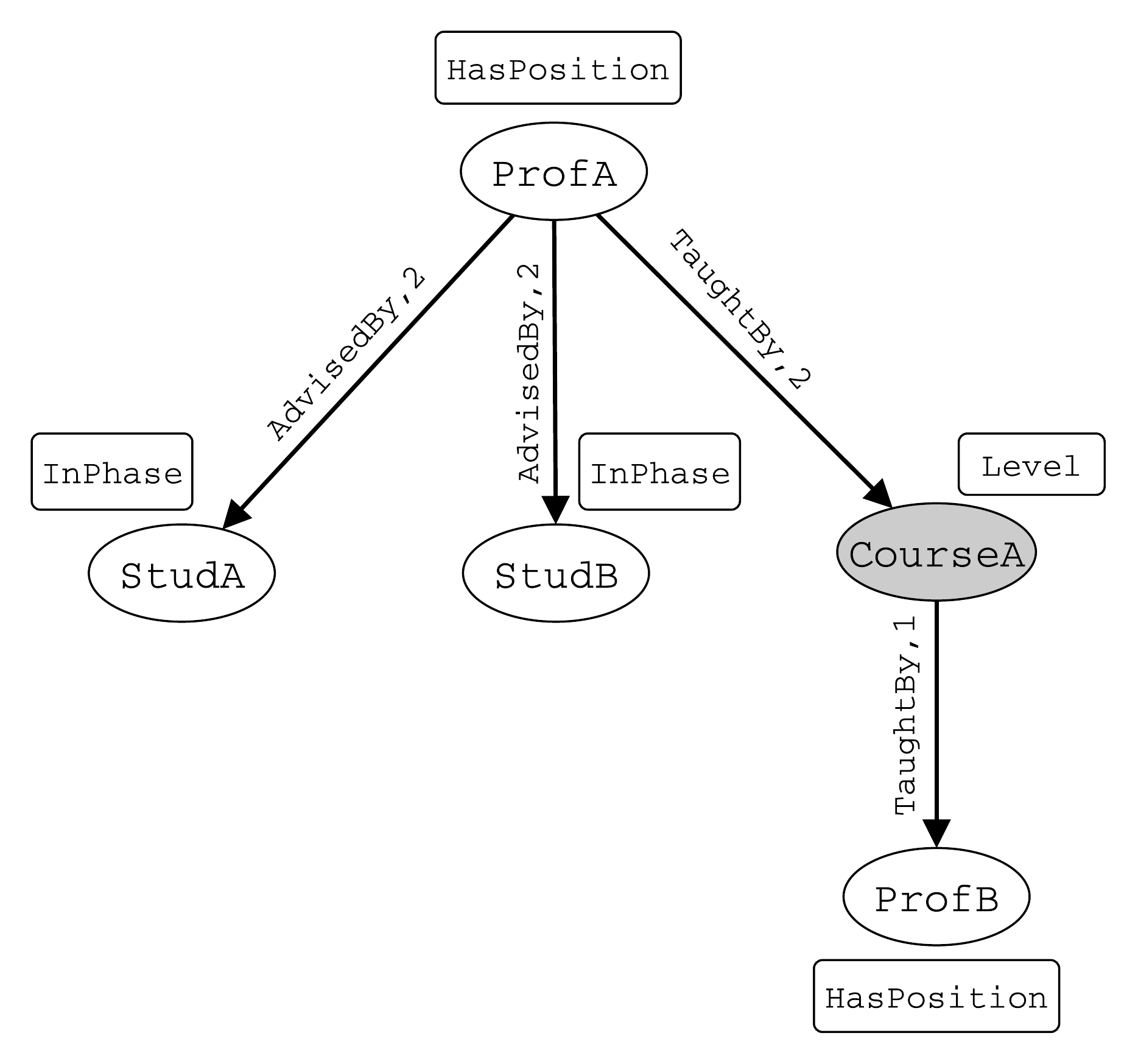}
    \end{minipage}
    \caption{A snapshot of a knowledge base (left) and the corresponding neighbourhood trees of \texttt{ProfA} entity (right). The knowledge base describes students, professors and courses they teach. Entities (people and courses) are represented with node, their attributes with rectangles and relationships with edges. Attribute values are left out for brevity. }
    \label{fig:NT}
\end{figure}

The central concept of CUR$^2$LED is a neighbourhood tree.
The neighbourhood tree is a rooted directed graph describing an instance, together with instances it relates to and their properties.
Viewing relational data as a hypergraph, the neighbourhood tree provides a summary of all path of a pre-defined length that originate in a particular vertex (see Figure~\ref{fig:NT}).

As instances are represented as neighbourhood trees, two instances are compared by comparing corresponding neighbourhood trees.
The authors introduce a versatile and declarative similarity measure \cite{Dumancic2017a} that analyses neighbourhood trees over multiple aspects by introducing the following \textit{core similarities}:
\begin{itemize}
	\item attribute similarity of root vertices
    \item attribute similarity of neighbouring vertices
    \item connectivity between root vertices
    \item similarity of vertex identities in a neighbourhood 
    \item similarity of edge types
\end{itemize}
Continuing the example in Figure~\ref{fig:NT}, person instances can be clustered based on their own attributes, which yields clusters of professors and students.
Clustering person instances based on the vertex identities in their neighbourhood yields clusters of \textit{research groups} -- a professor and his students.

These core similarities form  basic building blocks for a variety of similarity measure, all defined over neighbourhood trees.
The final similarity measure is a linear weighted combination of the core similarities.
Weights simply define a relative importance of core similarities in the final similarity measure.
The value assignments to the weights defines a \textit{similarity interpretation}.
For the details of core similarities and the similarity measure itself see \cite{Dumancic2017a}.

\section{Opening the black box of latent features}

Two ideas are central to CUR$^2$LED.
First, it learns latent features by clustering instances and their relationships.
Second, it uses multiple similarity interpretations (i.e., combinations of core similarities) to obtain a variety of features.
Both ideas are realised by means of neighbourhood trees.
Instances and relations are represented as (collections of) neighbourhood trees, while similarity interpretation is a result of core similarities which consider only certain parts of neighbourhood trees.

Latent features are learned by CUR$^2$LED through \textit{repeated clustering of instances and relations and alternating the similarity measure in each iteration}.
Each latent feature, corresponding to a cluster of instances, is associated with one latent predicate.
Truth instantiations of latent predicates reflect the cluster assignments, i.e., the instantiations of a latent predicate are true for instances that belong to the cluster; therefore, latent features are defined extensionally and lack an interpretable definition.
However, the intuitive specification of the similarity measure (and its core similarities) makes CUR$^2$LED a transparent method with a clear description which elements of neighbourhood trees make two instances similar.
Consequently, discovering the meaning of latent features is substantially easier than with the embedding approaches (and deep learning in general).

\subsection{Extracting the meaning of latent features}

Each latent feature corresponds to a cluster and the meaning of the features is reflected in the \textit{prototype} of the cluster.
To approximate the \textit{mean} or \textit{prototypical} neighbourhood tree, we search for the elements common to all neighbourhood trees forming a cluster.
These elements can be either attribute values, edge types or vertex identities.
The similarity interpretations used to obtain the cluster limits which elements are considered to be a part of a definition.
Moreover, neighbourhood trees \cite{Dumancic2017a} are compared by the relative frequencies of their elements, not the existence only.
Therefore, to find a mean neighbourhood tree and the meaning of a latent feature, we search for \textit{the elements with similar relative frequencies within each neighbourhood tree forming a cluster}.

\begin{figure}[t]
	\centering
    \includegraphics[scale=0.3]{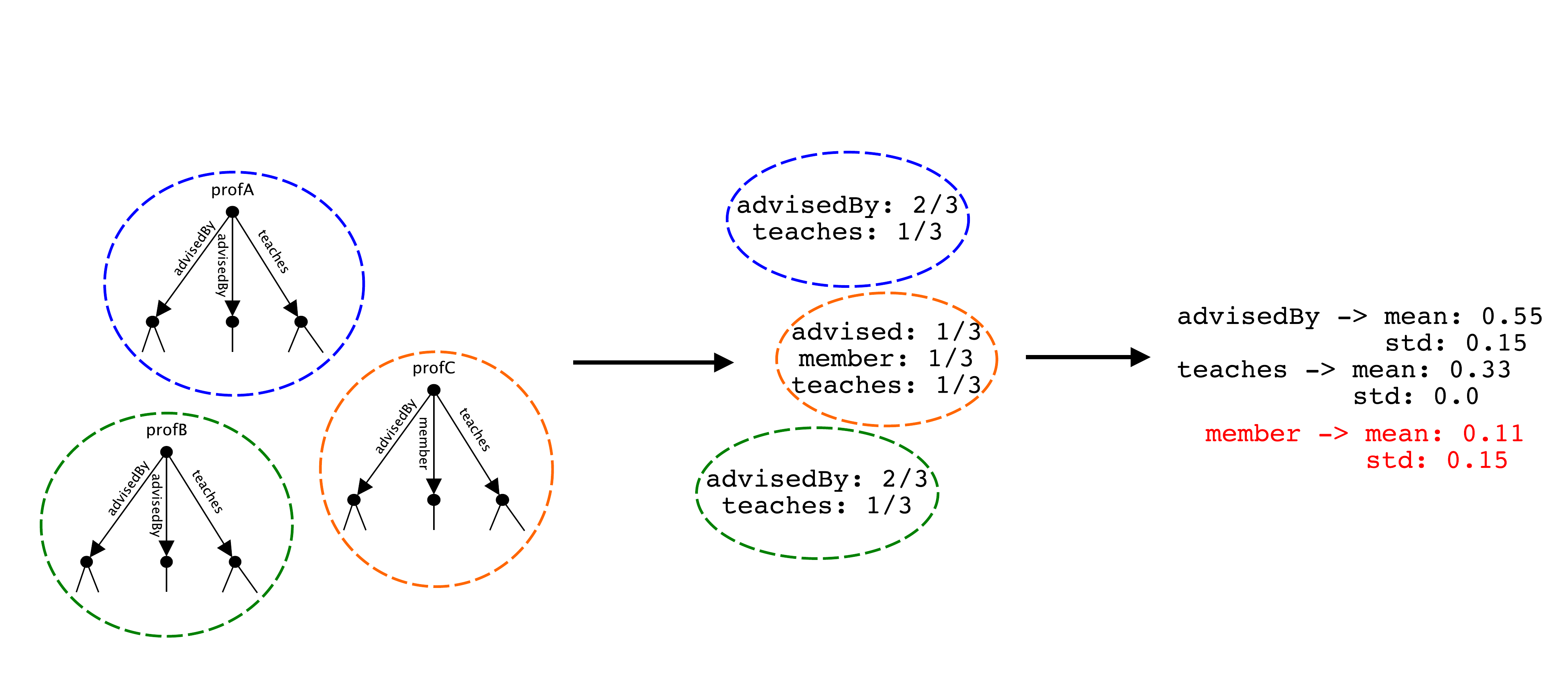}
    \caption{\textbf{Discovering the meaning of latent features by analysing their relations.} Properties that describe latent features are the ones that have similar relative frequency in all neighbourhood trees. Starting from a cluster of instances viewed as neighbourhood trees (left), the relative frequencies of elements are calculated for each neighbourhood tree (middle). Next, the mean and standard deviation of relative frequencies are calculated for each individual element within the cluster (right). Which elements \textit{explain} the latent features is decided with $\theta$-confidence. Setting $\theta$ to 0.3 identifies \texttt{advisedBy} and \texttt{teaches} as relevant elements (in black).}
    \label{fig:Process}
\end{figure}

To identify such elements, we proceed in three steps illustrated in Figure~\ref{fig:Process}.
\begin{enumerate}
	\item \textbf{Calculate the relative frequencies of all elements within each individual neighbourhood tree, per level and vertex type}.
    	In case of discrete attributes, that corresponds to a distribution of its values.
		In case of numerical attributes, we consider its mean value.
		In case of vertex identities and edge types, we simply look into their frequencies with respect to the depth in a neighbourhood tree.
        In the example in Figure~\ref{fig:Process}, the neighbourhood tree for \texttt{profA} contains two \texttt{advisedBy} relations, thus its frequency is $\frac{2}{3}$.
    \item \textbf{Calculate the mean and standard deviation of relative frequency for each element within a cluster}.
        In Figure~\ref{fig:Process}, the frequencies of the \texttt{advisedBy} elements in individual neighbourhood trees are $\frac{2}{3}, \frac{2}{3} \text{and} \frac{1}{3}$. Thus, its mean is $0.55$ with a standard deviation of $0.15$.
    \item \textbf{Select relevant elements.}
    	The final step involves a decision which elements should form a definition of a latent feature.
        Relevant elements are identified by a notion of \textit{$\theta$-confidence} which captures the allowed amount of variance in order to element to be relevant.
\end{enumerate}

\begin{definition}{\textbf{($\theta$-confidence)}}
An element with mean value $\mu$ and standard deviation $\sigma$  in a cluster,  is said to be $\theta$-confident if $\sigma \in [0, \theta \cdot \mu]$.
\end{definition}
 
In Figure~\ref{fig:Process}, setting $\theta$ to 0.3 makes \texttt{advisedBy} a $0.3$-confident element, because its standard deviation of 0.15 is within the range $[0, 0.3 \cdot 0.55] = [0, 0.165]$ specified by $\theta$.
In contrast, \texttt{member} is not a $0.3$-confident elements as its standard deviation is outside the range $[0, 0.3 \cdot 0.11] = [0, 0.0363]$.

The above-described procedure explains the latent features in terms of distribution of the elements in the neighbourhood of an instance, which has its pros and cons.
On the downside, this type of explanation does not conform to the standard first-order logic syntax common within relational learning.
Despite this reduced readability, these explanations are substantially more transparent and interpretable than the ones produced by the embeddings approaches.
However, an  benefit of this approach is that it increases the expressivity of a relational learner by extensionally defining  properties otherwise inexpressible in the first-order logic.

\subsection{Properties of latent spaces}

Latent features produced by CUR$^2$LED have proven useful in reducing the complexity of models and improving their performance.
However, no explanation was offered why that is the case.
In the second part of this work, we look into the properties of these latent representations and offer a partial explanation for their usefulness.
To answer this question we introduce the following properties: label entropy, sparsity and redundancy.

\textbf{Entropy and sparsity.}
Label entropy and sparsity serve as a proxy to a quantification of learning difficulty -- i.e., how difficult is it to learn a definition of the target concept.
Considering a particular predicate, label entropy reflects a \textit{purity} of its true groundings with respect to the provided labels.
Intuitively, if true groundings of predicates tend to predominantly focus on one particular label, we expect model learning to be easier.

Sparse representations, one of the cornerstones of deep learning \cite{Bengio:2013:RLR:2498740.2498889}, refer to a notion in which concepts are explained based on local (instead of global) properties of instance space.
Even though many properties might exist for a particular problem, sparse representations describe instances using only a small subset of those properties. 
Intuitively, a concept spread across a small number of local regions is expected to be easier to capture than a concept spread globally over an entire instance space.
Quantifying sparsity in relational data is a challenging task which can be approached from multiple directions -- either by analysing the number of true groundings or interaction between entities, for instance.
We adopt a simple definition: the number of true groundings of a predicate.

Label entropy and sparsity jointly describe a compelling property of data representation --  instances space is divided in many local regions that match labels well  and consequently make learning substantially easier.

\textbf{Redundancy.} 
A downside of CUR$^2$LED is the high number of  created features. 
Despite their proven usefulness, a high number of latent features enlarges the search space of a relational model and increases the difficulty of learning.
As similarity interpretations are provided by the user, it is possible that almost identical clusterings are obtained with different similarity interpretations.
Thus, if many of the features are redundant, removing them simplifies learning.
We measure the redundancy with the \textit{adjusted Rand index} (ARI) \cite{MoreyARI}, a standard measure for overlap between clusterings, and study its impact on the performance.

To evaluate the influence of redundant features, we modify CUR$^2$LED by adding an additional \textit{overlap parameter} $\alpha$.
Every time a new clustering is obtained, we check its overlap with the previously discovered clusterings using the ARI.
If the calculated value is bigger than $\alpha$, the clustering is rejected.

\section{Experiments and results}

We devise the experiments to answer the following questions:
\begin{itemize}
	\item[\textbf{(Q1)}] \textit{Are latent features created by CUR$^2$LED interpretable and do they capture sensible information?}
    \item[\textbf{(Q2)}] \textit{Do latent features that result in models of lower complexity and/or improved performance exhibit a lower label entropy compared to the original data representation?}
    \item[\textbf{(Q3)}] \textit{Are latent representation that improve the performance of a model sparser than the original data representations?}
    \item[\textbf{(Q4)}] \textit{To which extent are latent features redundant?}
\end{itemize}

\subsection{Datasets and setup}

The results  obtained in \cite{Dumancic2017} can be divided in three categories.
The first category contains the IMDB and UWCSE datasets; these datasets present easy relational learning tasks in which the original data representation is sufficient for almost perfect performance.
The main benefit of latent representations for these tasks was the reduction of model complexity.
The second category includes the TerroristAttack dataset., in which the main benefit of latent representation was the reduction of complexity, but not the performance.
The third category involves the Hepatitis, Mutagenesis and WebKB datasets.
These tasks benefited from latent representations in both performance and reduced model complexity.
That is especially true for the Hepatitis and WebKB datasets on which the performance was improved by a large margin.

We take a representative task from each of the categories.
Precisely, we use IMDB, UWCSE, Hepatitis and TerroristAttack datasets in our experiments.
Both IMDB and UWCSE datasets were included as they are easy to understand without the domain knowledge, and thus useful for analysing the interpretability of relational latent features.
As for the parameters of latent representation, we take the best parameters on individual datasets selected by the model selection procedure in \cite{Dumancic2017}.
When analysing the interpretability, we set $\theta$ to $0.3$.

When evaluating the redundancy, we create latent representations by setting the $\alpha$ to the following values: $\{0.9, 0.8, 0.7, 0.6, 0.5\}$.
We then learn a relational decision tree TILDE \cite{Blockeel1998285} on the obtained representation and compare accuracies, the number of created features and the number of facts.

\subsection{Interpretability}

\begin{figure}[t]
	\centering
    \includegraphics[scale=0.3]{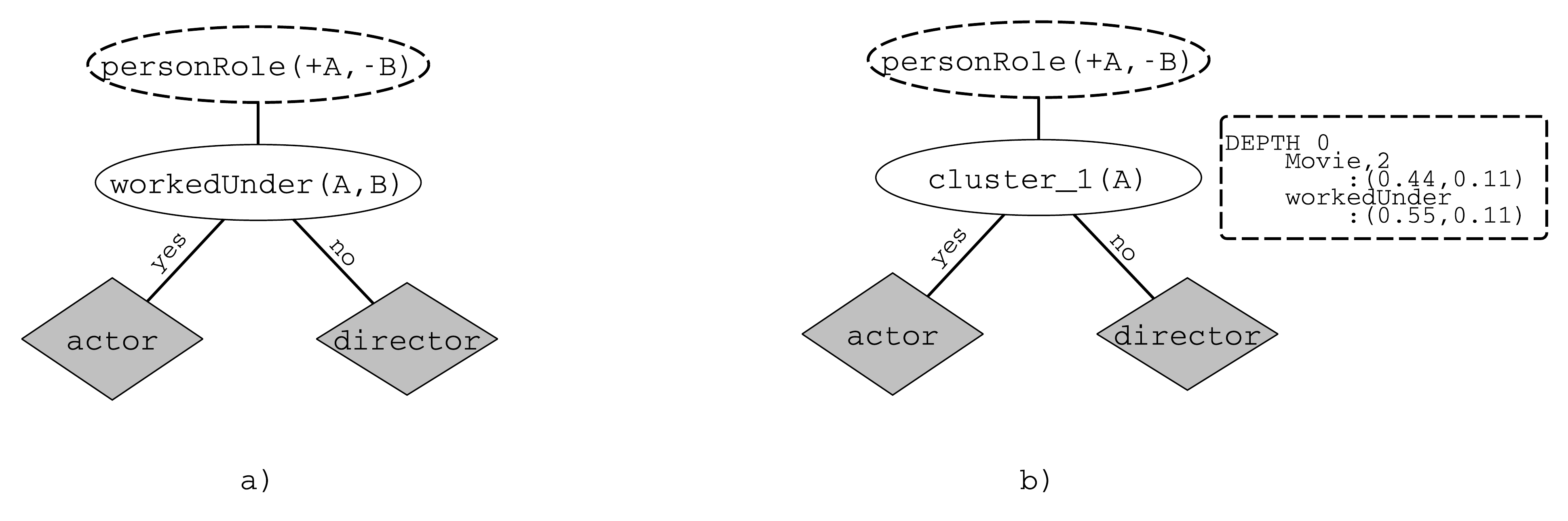}
    \caption{Relational decision trees learned on the original (left) and latent (right) data representation of the IMDB dataset. The dashed ellipse indicates the target predicate and its arguments. The first argument, marked \texttt{A} and declared as input (\texttt{+}), denotes a person. The second argument, marked \texttt{B} and declared as output (\texttt{-}), states the label of the instance given by \texttt{A}. The values in the leaves of the decision trees are assignments to \texttt{B}. The dashed rectangle describes the latent feature -- for each level of the \textit{mean neighbourhood tree}, $\theta$-confident elements are listed with the mean and standard deviation.  }
    \label{fig:IMDBtree}
\end{figure}

To illustrate the interpretability of relational features, we show examples of latent features created for different datasets.
We show that the relational decision trees  learned on both original and latent representations.
The explanations of latent features are provided as well.

Figure~\ref{fig:IMDBtree} shows the decision trees learned on the IMDB dataset.
The task is to distinguish between actors and directors -- this is a simple relational learning task and both original and latent decision tree achieve the perfect performance with only a single node.
Even though latent representation does not seem beneficial in this particular case, it is interesting to see that the selected latent feature captures the same information as the decision tree learned on the original data -- person instances in \texttt{cluster\_1} are the ones that have a relationship with \texttt{movie} instances, and have worked under another person (a director).

\begin{figure}[t]
	\centering\includegraphics[scale=0.28]{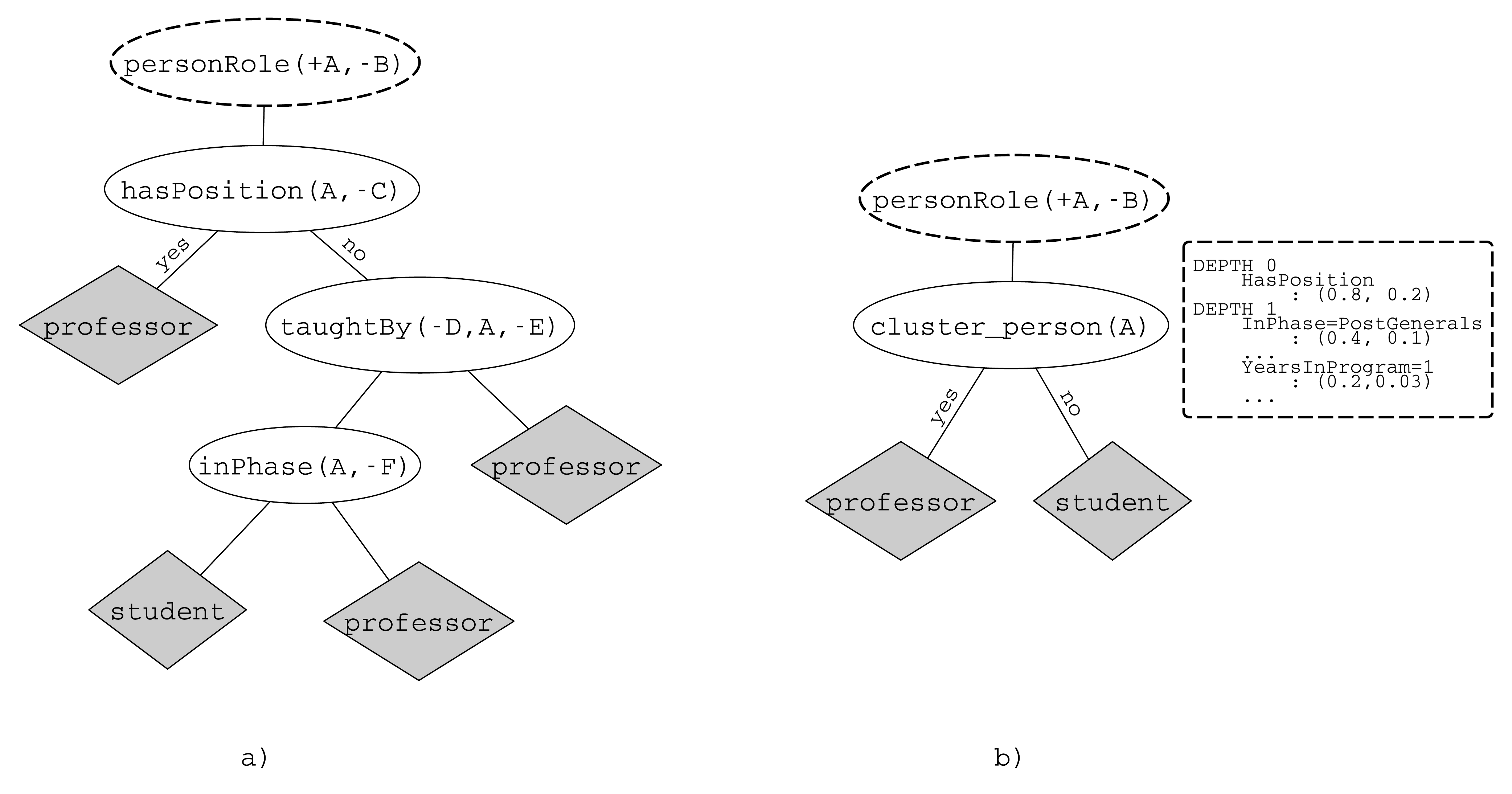}
    \caption{Relational decision trees learned on the original (left) and latent (right) representations of the UWCSE dataset. The elements have the same meanings as in Figure~\ref{fig:IMDBtree}. }
    \label{fig:UWCSEtree}
\end{figure}

Figure~\ref{fig:UWCSEtree} shows the decision trees for the UWCSE dataset, which benefit from the latent features.
Despite the simplicity of distinguishing students from professors, the decision tree learned on the latent features is more compact and has only a single node whereas the decision tree learned on the original features consists of three nodes.
The latent feature here again captures similar knowledge as the original decision tree but expressed in a simpler manner -- professor is someone who either has a position at the faculty, or is connected to people who are currently in a certain phase of a study program and have been in the program for a certain number of years.

What is particularly interesting about the examples above is that, even though the latent features are created in an unsupervised manner, they match the provided label very well.
Moreover, they seem to almost perfectly capture the labelled information as only a few features are needed to outperform the decision tree learned on the original data representation.
This observation shows that CUR$^2$LED is indeed capturing sensible knowledge in the latent space.

Both aforementioned examples are easy to understand and interpret without an extensive domain knowledge.
The other tasks that have benefited more from the latent features are substantially more difficult to understand.
For instance, the latent features created from the Mutagenesis dataset reduce the complexity of the relational decision tree from 27 to only 3 nodes, while improving the accuracy for 4 \%.
Similarly, on the Hepatitis dataset the latent features reduced the complexity of a decision tree from 22 nodes down to 5, improving the accuracy for 11 \%.
Because these examples require an extensive knowledge to interpret them, we leave them out from this work.

\subsection{Properties of latent spaces}

\begin{figure}[t]
	\centering
    \begin{minipage}{0.24\textwidth}
    \centering
    \includegraphics[scale=0.32]{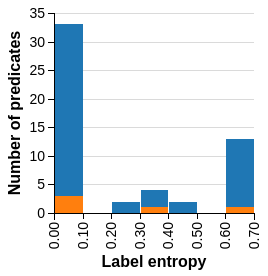}
    
    a) IMDB    
    \end{minipage}
    \begin{minipage}{0.24\textwidth}
    \centering
    \includegraphics[scale=0.32]{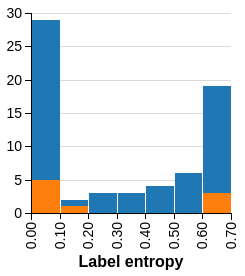}
    
    b) UWCSE
    \end{minipage}
    \begin{minipage}{0.24\textwidth}
    \centering
    \includegraphics[scale=0.32]{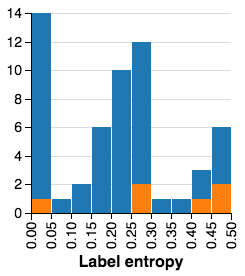}
    
    c) Hepatitis
    \end{minipage}
    \begin{minipage}{0.24\textwidth}
    \centering
    \includegraphics[scale=0.3]{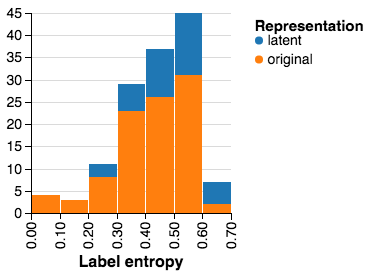}
    
    d) TerroristAttacks
    \end{minipage}
    \caption{The histogram of label entropy values for latent and original representations (indicated by the colour).}
    \label{fig:Entropy}
\end{figure}

\textbf{Label entropy.}
Figure~\ref{fig:Entropy} summarizes the label entropy for each dataset.
In all cases where representation learning proved helpful (i.e., IMDB, UWCSE, Hepatitis), latent representations have a substantially larger number of predicates with low label entropy compared to the original data representation. 
The latent representation for the TerroristAttack datasets, however, shows a different behaviour in which latent features with high entropy dominate the representation.
These results agree with the expectation that a high number of low entropy features makes learning easier.
However, not all latent features have low label entropy.
This is expected, as the labels are not considered during learning of latent features.
It also does not pose a problem -- these latent features are less consistent with the one particular task, but it easily might be the case that those features are useful for a different task.

\begin{figure}[t]
	\centering
    \begin{minipage}{0.24\textwidth}
    \centering
    \includegraphics[scale=0.3]{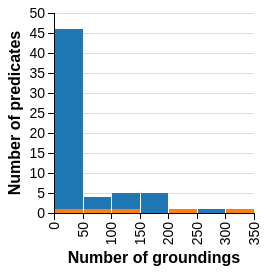}
    
    a) IMDB    
    \end{minipage}
    \begin{minipage}{0.24\textwidth}
    \centering
    \includegraphics[scale=0.3]{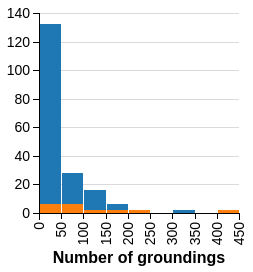}
    
    b) UWCSE
    \end{minipage}
    \begin{minipage}{0.24\textwidth}
    \centering
    \includegraphics[scale=0.3]{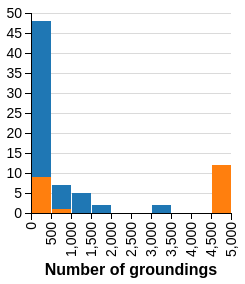}
    
    c) Hepatitis
    \end{minipage}
    \begin{minipage}{0.24\textwidth}
    \centering
    \includegraphics[scale=0.3]{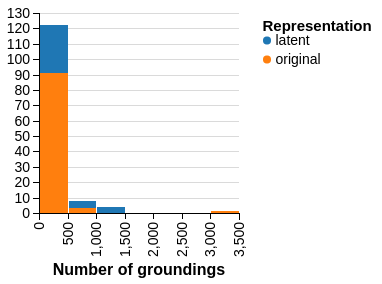}
    
    d) TerroristAttacks
    \end{minipage}
    \caption{Histogram of the number of true groundings of predicate for latent on original representations (indicated by the colour).}
    \label{fig:Sparsity}
\end{figure}

\textbf{Sparsity.}
Figure~\ref{fig:Sparsity} summarizes the number of groundings, i.e., the sparsity.
The distribution of the number of true groundings in the latent representations (where latent features are beneficial) is heavily skewed towards a small number of groundings, in contrast with the original representation.
That is especially the case with the Hepatitis dataset, which profits the most from the latent features.
The exception to this behaviour is again the TerroristAttack dataset in which the original representation already is very sparse.
These results indicates that latent features indeed describe smaller groups of instances and their local properties, instead of global properties of all instances.

\begin{figure}[t]
	\centering
    \begin{minipage}{0.24\textwidth}
    \centering
    \includegraphics[scale=0.3]{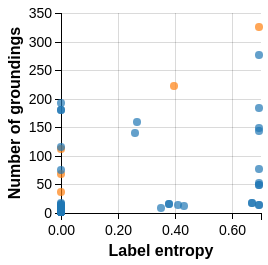}
    
    a) IMDB    
    \end{minipage}
    \begin{minipage}{0.24\textwidth}
    \centering
    \includegraphics[scale=0.3]{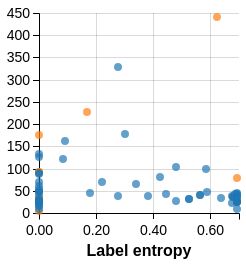}
    
    b) UWCSE
    \end{minipage}
    \begin{minipage}{0.24\textwidth}
    \centering
    \includegraphics[scale=0.3]{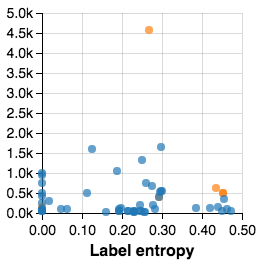}
    
    c) Hepatitis
    \end{minipage}
    \begin{minipage}{0.24\textwidth}
    \centering
    \includegraphics[scale=0.3]{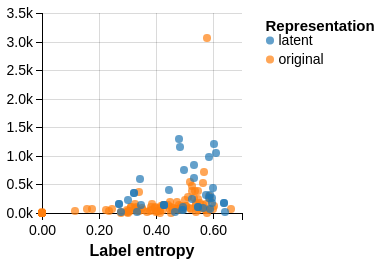}
    
    d) TerroristAttacks
    \end{minipage}
    \caption{Contrasting the label entropy of predicates (horizontal axis) and the number of true groundings (vertical axis). Whether a predicate comes from the original or latent representation is indicated by the colour.}
    \label{fig:EntropyVsSparsity}
\end{figure}

\textbf{Connecting label entropy and sparsity.}
A potential explanation of the above discussed results might be that many latent features capture a very small number of instances (e.g., 1 or 2) which consequently leads to a large number of features with low label entropy.
Such features would largely be useless as they make generalization very difficult.
To verify that this is not the case, Figure~\ref{fig:EntropyVsSparsity} plots the label entropy versus the number of groundings of a predicate.
If latent features of low label entropy would indeed capture only a small number of instances, many points would be condensed in the bottom left corner of the plot.
However, that is not the case -- many latent predicates with low label entropy actually have a number of groundings comparable to the predicates in the original representation.
The exception to this is again the TerroristAttacks dataset.

These results jointly point to the following conclusion: \textit{latent features successfully identify local regions in the instance space that match well with the provided labels}.
As a consequence, these local regions are easier to capture and represent.

\begin{figure}[t]
	\centering
    \begin{minipage}{0.24\textwidth}
    \centering
    \includegraphics[scale=0.4]{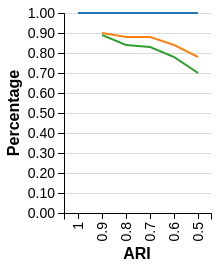}
    
    a) IMDB    
    \end{minipage}
    \begin{minipage}{0.24\textwidth}
    \centering
    \includegraphics[scale=0.4]{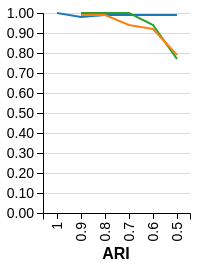}
    
    b) UWCSE
    \end{minipage}
    \begin{minipage}{0.24\textwidth}
    \centering
    \includegraphics[scale=0.4]{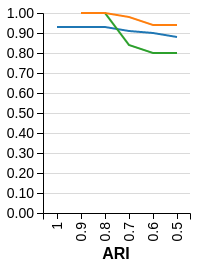}
    
    c) Hepatitis
    \end{minipage}
    \begin{minipage}{0.24\textwidth}
    \centering
    \includegraphics[scale=0.4]{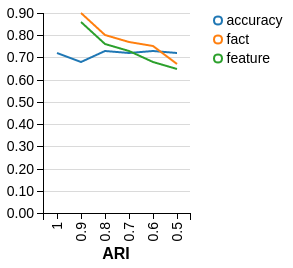}
    
    d) TerroristAttacks
    \end{minipage}
    \caption{\textbf{Redundancy of features in latent representations.} The accuracy (blue line), the number of latent features (green line) and the number of facts (= a sum of true groundings of all predicates, orange line) are reported for varying values of $\alpha$. For the accuracy, the percentage of correctly classified examples is reported. For the number of features and facts, we report the ratio between the number of features/facts in the latent representation obtained with a specific value for $\alpha$ and the number of features/facts in the latent representation with $\alpha=1.0$.  }
    \label{fig:Redundancy}
\end{figure}

\textbf{Redundancy.}
Figure~\ref{fig:Redundancy} summarizes the influence of $\alpha$ on the accuracy and the number of latent features.
These results show that the performance of the classifier is not affected by removing features based on the overlap of clusterings they define.
The performance of TILDE remains approximately the same, whereas the number of latent features is reduced by 20 to 30 \%.
As the number of features is directly related to the size of the search space of relational model (and thus the complexity of learning), this is an encouraging result indicating that the size of the search space can be naively reduced without sacrificing the performance.

\section{Conclusion}

In this work we closely inspect the properties of latent representations for relational data.
We focus on relational latent representations created by clustering both instances and relations among them, introduced by CUR$^2$LED \cite{Dumancic2017}.
The first property we analyse is the interpretability of latent features.
We introduce a simple method to explain the meaning of latent features, and show that they capture interesting and sensible properties.
Second, we identify two properties of these latent representation that partially explain their usefulness -- namely, the label entropy and sparsity.
Using these two properties, we show that obtained latent features identify local regions in instance space that match well with the labels.
Consequently, this explains why predictive model learned from latent features are less complex and often perform better than the model learned from the original features.
Third, we show that that latent features tend to be redundant, and that 20 to 30 \% of latent features can be discarded without sacrificing the performance of the classifier.
This consequently reduces the search space for the relational models, and simplifies learning.

\subsection*{Acknowledgements}
This research is supported by Research Fund KU Leuven (GOA/13/010).

\bibliographystyle{unsrt}
\bibliography{bibliography}

\end{document}